\documentclass[lettersize,journal]{IEEEtran}
\usepackage{amsmath,amsfonts}
\usepackage{algorithmic}
\usepackage{algorithm}
\usepackage{array}
\usepackage[caption=false,font=normalsize,labelfont=sf,textfont=sf]{subfig}
\usepackage{textcomp}
\usepackage{stfloats}
\usepackage{url}
\usepackage{verbatim}
\usepackage{graphicx}
\usepackage{cite}

\usepackage[table,xcdraw]{xcolor}
\usepackage{colortbl}
\usepackage{booktabs}
\usepackage{multirow}

\usepackage{algorithm}
\usepackage{algorithmic}
\usepackage{amsmath}
\usepackage{amssymb}
\usepackage{tabularx}

\newcommand{\ie}{\textit{i.e.}}

\begin{document}

\title{Learning to Rank Pre-trained Vision-Language Models for Downstream Tasks}
\author{Yuhe Ding, Bo Jiang, Aihua Zheng, Qin Xu and Jian Liang

\thanks{Yuhe Ding, Bo Jiang, and Qin Xu are with the School of Computer Science and Technology, Anhui University. E-mail: madao3c@foxmail.com; jiangbo@ahu.edu.cn; xuqin@ahu.edu.cn.}
\thanks{Aihua Zheng is with the School of Artificial Intelligence, Anhui University. E-mail: ahzheng214@foxmail.com.}
\thanks{Jian Liang is with the New Laboratory of Pattern Recognition, Institute of Automation, Chinese Academy of Sciences. E-mail: liangjian92@gmail.com.}
\thanks{Bo Jiang and Jian Liang are the corresponding authors.}
}

\markboth{Journal of \LaTeX\ Class Files,~Vol.~14, No.~8, August~2021}%
{Shell \MakeLowercase{\textit{et al.}}: A Sample Article Using IEEEtran.cls for IEEE Journals}


\maketitle

\begin{abstract}
Vision language models (VLMs) like CLIP show stellar zero-shot capability on classification benchmarks.
However, selecting the VLM with the highest performance on the unlabeled downstream task is non-trivial.
Existing VLM selection methods focus on the class-name-only setting, relying on a supervised large-scale dataset and large language models, which may not be accessible or feasible during deployment.
This paper introduces the problem of \textbf{unsupervised vision-language model selection}, where only unsupervised downstream datasets are available, with no additional information provided.
To solve this problem, we propose a method termed Visual-tExtual Graph Alignment (VEGA), to select VLMs without any annotations by measuring the alignment of the VLM between the two modalities on the downstream task.
VEGA is motivated by the pretraining paradigm of VLMs, which aligns features with the same semantics from the visual and textual modalities, thereby mapping both modalities into a shared representation space.
Specifically, we first construct two graphs on the vision and textual features, respectively. 
VEGA is then defined as the overall similarity between the visual and textual graphs at both node and edge levels.
Extensive experiments across three different benchmarks, covering a variety of application scenarios and downstream datasets, demonstrate that VEGA consistently provides reliable and accurate estimates of VLMs' performance on unlabeled downstream tasks. 
\end{abstract}


\section{Introduction}
Vision language models (VLMs) like CLIP \cite{radford2021learning}, ALIGN \cite{jia2021scaling} and SigLIP \cite{zhai2023sigmoid}, are transforming the technological and academic landscape with their unprecedented performance and the broad range of viable applications \cite{peng2023sgva, xiao2023clip, yang2023effective, yang2023neural}.
The most impressive capability of VLMs is their applications in zero-shot classification tasks.
With just the class names, VLMs can be easily applied to any downstream task.
However, identifying which VLM has the highest downstream performance is non-trivial, as labels are unavailable when deployed in real-world scenarios.

\begin{figure}
    \centering
    \includegraphics[width=1.\linewidth]{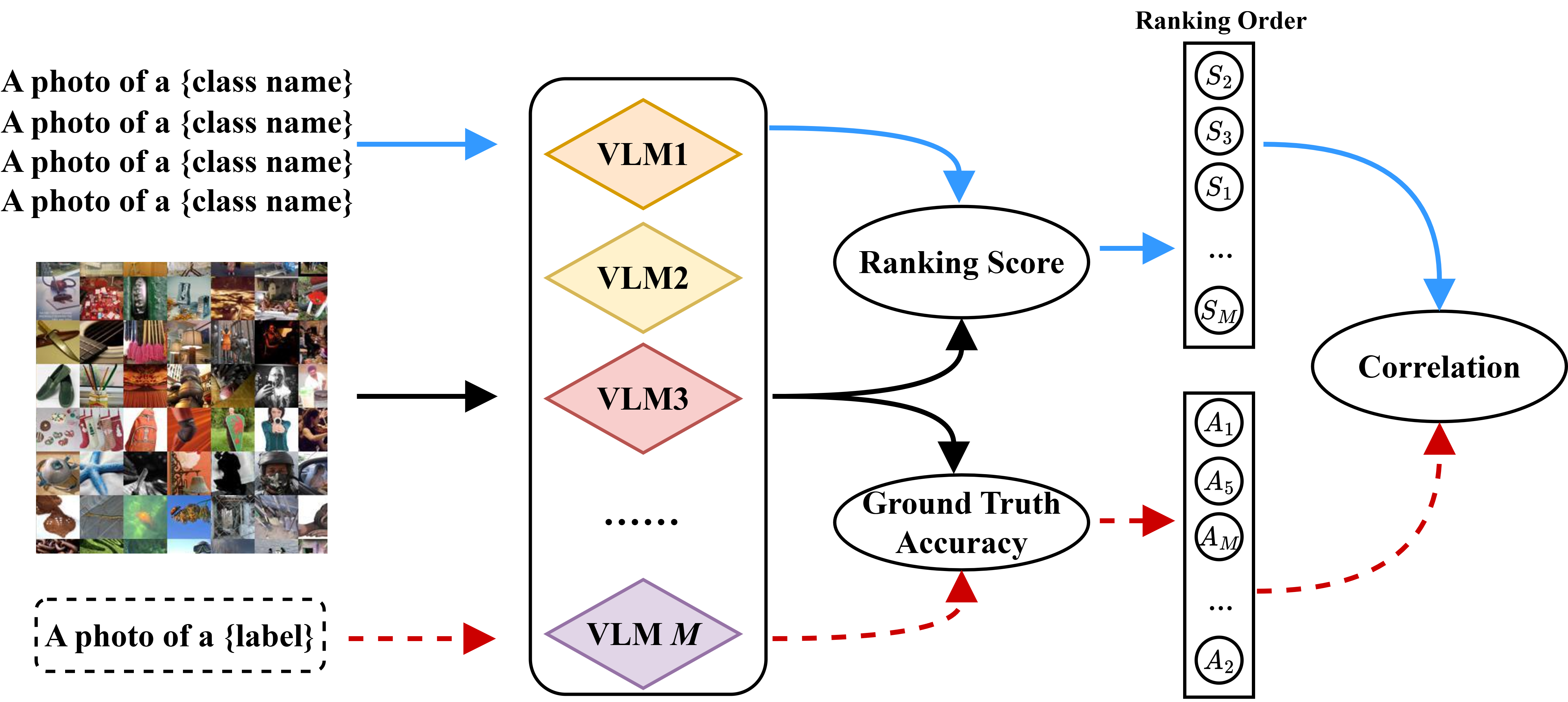}
    \caption{Paradigm of unsupervised vision language model selection, where only unsupervised downstream datasets are available, with no additional information provided. The goal is to develop a method that computes a score for each VLM, which is highly correlated with the unseen ground truth accuracy.}
    \label{fig:paradigm}
\end{figure}

Recently, language-only vision language model selection (LOVM) \cite{zohar2024lovm, yi2024bridge}, which selects a VLM for the downstream dataset with only class names, has garnered attention.
LOVM methods usually leverage the zero-shot classification accuracy on a large-scale dataset with annotations such as ImageNet \cite{russakovsky2015imagenet} as a baseline, and additionally introduce large language models (LLMs) \cite{openai2024gpt4technicalreport} to generate captions and synonyms for these class names.
However, the prediction results are sensitive to the quality of the content generated by LLM, and calling the LLM API can also be quite time-consuming and costly.
Besides, a dataset with annotations is not always available during deployment.
A viable solution is to use unsupervised downstream datasets along with the corresponding class names, which are readily accessible to downstream users in deployment scenarios.
Some methods tailored for traditional convolution neural network models \cite{garg2022leveraging, deng2021labels, yu2022predicting} also consider this problem.
They typically predict downstream performance (also known as generalization performance or out-of-distribution performance) by measuring the distribution divergence between the training and downstream datasets.
While this straightforward idea has been demonstrated to be applicable to VLMs \cite{fang2022data, mayilvahanan2023does}, implementing these methods directly on VLMs remains challenging.
The reason is that training data is often difficult for downstream users to access, either due to its huge size or restrictions imposed by privacy and commercial considerations.
Different from the two settings mentioned above, unsupervised vision language model selection aims to select VLMs using only the unlabeled target dataset.
The paradigm is shown in Fig. \ref{fig:paradigm}, and this setting is practical and can eliminate the dependency on training datasets and LLMs presented in existing methods. 

To solve this problem, we propose Visual-tExtual Graph Alignment (VEGA), a new method to evaluate the downstream performance of pre-trained VLMs with the corresponding unsupervised downstream dataset.
VEGA is motivated by the pretraining paradigm of VLMs, which aligns features with the same semantics from the visual and textual modalities, thereby mapping both modalities into a shared representation space.
In a well-trained cross-modality features space, visual features should be tightly clustered around the corresponding textual features \cite{radford2021learning}.
This phenomenon leads to a straightforward intuition: the more similar the structures of the class feature distributions for the two modalities, the easier it becomes to match the images to their corresponding class names.
We model the structures of the class distributions in the two modalities as a fully connected visual graph and textual graph, respectively.
Both graphs have the same number of nodes, with each node representing a class and edges representing the distances between connected classes.
Specifically, the node and edge of the textual graph are simply defined as the textual feature of class names and the cosine distance, respectively.
In the visual graph, nodes correspond to clusters of visual features of images, which are closest to the corresponding class name features, and edges represent the Bhattacharyya distances between the nodes.
VEGA represents the similarity between the two graphs by combining both node and edge level similarity together. 
Specifically, node similarity is the average distance between image features in a visual node and the corresponding textual node. 
Edge similarity is the Pearson correlation coefficient between the edge matrices, which can eliminate the impact of scale.
VLMs with a higher VEGA score are more likely to achieve better downstream performance.

We conduct extensive experiments on three practical application scenarios of VLM performance prediction, \ie, VLMs from the CLIP family, VLMs from various pre-training algorithms, and the combination of VLM and prompt template. 
The results validate that VEGA is a reliable downstream performance indicator under various practical scenarios.
The contributions of this study can be summarized as follows,
\begin{itemize}
    \item We introduce a new problem setting that is practical for downstream users: unsupervised vision language model selection, where class names and unlabeled downstream datasets are available.
    \item We propose a novel method termed Visual-tExtual Graph Alignment (VEGA), which measures the similarity between the well-designed class distribution graphs of the visual and textual modalities, serving as an estimator of VLM zero-shot classification performance.
    \item We provide three benchmarks for this new setting, involving performance prediction on VLMs from the CLIP family, VLMs from various pre-training algorithms, and the combination of VLM and prompt template. Superior results validate that VEGA is a reliable unsupervised indicator of VLM downstream performance.
\end{itemize}

\section{Related Work}
\subsection{Model Selection}
Model selection, a core challenge in transfer learning, focuses on ranking available pre-trained models to identify the one best suited for a given target task \cite{yu2024survey, ding2024model, garrido2023rankme}.
Model selection can be divided into several popular topics based on the different goals of the target task.
Transferability estimation \cite{ding2024model, guo2023identifying, gholami2023etran} aims to maximize the accuracy of the target task after supervised fine-tuning.
The difficulty lies in how to select a model using a supervised target dataset without the need for fine-tuning or a small amount of fine-tuning.
Out-of-distribution (OOD) error prediction \cite{yu2024survey, lu2023predicting} focuses on evaluating a model's ability to maintain robust performance when presented with data that deviates from its training distribution. 
These approaches involve using a test set specifically designed to include OOD data, allowing for an assessment of the model's generalization capacity under challenging, unseen conditions. 
Unlike traditional transfer learning approaches that require fine-tuning on downstream tasks, OOD error prediction remains within the same task framework, aiming to measure how well the model adapts to variations in data distributions without additional training. 
This evaluation provides insights into the model's resilience and reliability in real-world scenarios where data distribution shifts are inevitable.
Model validation \cite{mosteller1977data} is a crucial step in the machine learning workflow, enabling the evaluation and comparison of different training checkpoints to identify the most effective model. 
In supervised validation \cite{mosteller1977data}, a labeled validation set is used to measure performance and select the model with the best validation metrics, ensuring its ability to generalize to unseen data.
In contrast, unsupervised validation \cite{saito2021tune, sugiyama2007covariate, you2019towards} addresses scenarios where labeled validation data is unavailable. It leverages the unlabeled test set or proxy metrics to assess model performance, providing an alternative means for model selection in settings where labeling data is challenging or infeasible.

\subsection{Vision-language Model Selection}
LOVM \cite{zohar2024lovm} introduces a new setting termed language-only vision language model selection task, where methods are expected to perform both model selection and performance prediction based solely on a text description of the desired downstream application.
LOVM generates a caption dataset and a synonym dataset and then calculates several statistic scores on these text datasets.
This is an interesting setting and is reasonable in cases where data is extremely limited.
However, LOVM relies on large language models (LLMs) \cite{openai2024gpt4technicalreport} to generate a substantial number of captions and synonyms for these class names.
The prediction results are sensitive to the quality of the content generated by the LLM, and calling the LLM API can also be quite time-consuming and costly.
Besides, some recent studies \cite{fang2022data, radford2021learning, mayilvahanan2023does} find that the generalization performance has a high correlation with train-test set similarity.
They design various methods to measure train-test set similarity.
For downstream users, the training set is difficult to obtain, while the downstream dataset is usually available.
Therefore, developing a downstream performance evaluation method for vision language models with a downstream unsupervised dataset is practical.

\subsection{Generalization Performance Prediction.} 
As the rapid proliferation of generalization algorithms such as domain generalization \cite{yu2024survey}, distributionally robust optimization \cite{namkoong2016stochastic}, invariant learning \cite{arjovsky2019invariant} and stable learning \cite{shen2020stable}, etc, evaluating their ability under possible distribution shift scenarios becomes increasingly critical for a downstream application.
Existing generalization performance prediction methods are divided into several types.
Confidence-based methods \cite{garg2022leveraging, hendrycks17baseline} are based on the intuition that the performance of models is related to their prediction confidence.
Discrepancy-based methods measure the distribution discrepancy between the training and test sets, with the aid of some classical metrics such as Frechec Distance \cite{deng2021labels} or well-designed methods such as projection norm \cite{yu2022predicting}.
Consistency-based methods measure the consistency of models under diverse scenarios and tasks \cite{deng2021does,baek2022agreement}.
Actually, most generalization performance methods rely on the training data (also known as in-distribution, known distribution, source data, etc). 
However, for VLMs, the training data is huge, and some of it may be inaccessible due to privacy or commercial reasons, making it challenging to apply these methods directly to the performance prediction task of VLMs.
We select four representative methods that do not strictly rely on training data and compare them in our experiments.

\section{Preliminary}
In this section, we formally introduce the setting of unsupervised vision language model selection.

\subsection{Zero-shot Classification of Vision Language Models}
We denote the candidate VLMs as $\{v_m = (\phi_m,\xi_m )\}_{m=1}^{M}$, where $\phi_m$ and $\xi_m$ notate the visual encoder and textual encoder of $m$-th VLM, respectively;
$X=\{x_i\}_{i=1}^{N}$ denotes the unlabeled downstream dataset, where $N$ is the number of images.
$C=\{c_k\}_{k=1}^K$ represents the class names, \ie, label space, where $K$ is the number of classes. 
Zero-shot classification with VLMs involves encoding both image and text prompts (\textit{e.g.}, ``a photo of a \{class name\}") into feature vectors. 
An image is classified by selecting the class whose textual feature has the highest cosine similarity to the image's feature vector,
\begin{equation}\label{eq:yhat}
     \hat{y_i} = \mathop{argmax}\limits_k(cos(\xi_m(\Tilde{c}_k), \phi_m(x_i))),
\end{equation}
where $cos(\cdot)$ is the cosine similarity, $\Tilde{c}_k$ is the text prompt of the class name $c_k$, $y_i$ is the real label of $x_i$; $\phi_m(x_i) \in \mathbb{R}^{D}$ and $\xi_m(\Tilde{c_k}) \in \mathbb{R}^{D\times K}$ denote the visual feature and textual feature respectively, $D$ is the dimension of features.
It is worth noting that, text prompts also play a crucial role when employing VLMs for zero-shot classification. 
Selecting an appropriate prompt template is essential, as it significantly impacts the effectiveness of zero-shot classification.
Notate the templates as $\{\sigma_p\}_{p=1}^P$, $P$ is the number of candidate templates, the text prompts $\Tilde{c}_k$ are defined as $\Tilde{c}_k = \sigma_i(c_k)$.
For different VLMs, the optimal template is not necessarily the same, so it is equally important to choose a suitable combination of VLM and template.

\subsection{Vision Language Model Selection}
A large number of VLMs have emerged in recent years. 
There are dozens of different model architectures in the CLIP \cite{radford2021learning, peng2023sgva} family alone, and diverse pre-training algorithms \cite{zhai2023sigmoid,li2022blip} also flourished.
Vision language model selection \cite{zohar2024lovm,yi2024bridge} aims to select a model for downstream datasets with the highest zero-shot classification accuracy. 
Formally, a VLM selection algorithm $h$ aims to calculate a score $s_m$ for each VLM $v_m=(\phi_m,\xi_m)$,
\begin{equation}
    s_m = h(v_m) = h(\phi_m,\xi_m),
\end{equation}
$s_m$ is highly-correlated with the zero-shot performance $a_m = \frac{1}{N}{\sum}_{i=1}^{N} \mathbb{I}(\hat{y_i} = y_i)$, where $\hat{y}$ is the defined in Eq. (\ref{eq:yhat}), and $y_i$ is the real label for each unlabeled image.

\noindent
\textbf{Language Only VLM Selection (LOVM).}
Existing methods focus on language-only VLM selection (LOVM) \cite{zohar2024lovm}, where only meta information, \ie, class names, are available,
\begin{equation}
    s_m = h_{LOVM}(v_m|C_d),
\end{equation}
where $C_d$ is the class name of dataset $d$.
As information is scarce, LOVM introduces the large language model (LLM) \cite{openai2024gpt4technicalreport}, which is important prior knowledge to this setting.
LLM generates many probable image captions, which could be encoded using the different VLM text encoders, producing the corresponding text embeddings, which are treated as image proxies. 
Existing work \cite{zohar2024lovm} introduces the accuracy of the candidate model on ImageNet \cite{russakovsky2015imagenet} (INB) as a baseline, and additionally proposes the text classification score (LOVM-C) and dataset granularity score (LOVM-G).
INB is a strong baseline, and its computation requires the full ImageNet dataset and its labels, which is not available in most real-world situations.

\noindent
\textbf{Unsupervised VLM Selection (UVMS).}
We focus on the Unsupervised VLM selection (UVMS) problem, where the unsupervised downstream data and the class names are available,
\begin{equation}
    s_m = h_{UVMS}(\phi_m, \xi_m | C, X).
\end{equation}
Due to the scarcity of supervision information, LOVM needs to introduce large-scale supervised datasets, \ie, ImageNet, and large language models. 
Our UVMS setting strictly requires only unsupervised downstream data to be available. 
This approach is more practical because we always have the test data during deployment, while the availability of a supervised dataset and LLMs is not guaranteed.

\noindent
\textbf{Evaluation of UVMS task.}
To evaluate the performance of the UVMS method comprehensively, we introduce four commonly used metrics in model evaluation methods \cite{zohar2024lovm, saito2021tune}.
Specifically, given the ground truth, \ie, the zero-shot classification accuracy $\mathcal{A} = \{a_m\}_{m=1}^M$ of the candidate models on the target dataset, and the predicted scores $\mathcal{S} = \{s_m\}_{m=1}^M$ of the candidate models, the metrics are introduced as follows:

\begin{itemize}
\item \textbf{Top-5 Recall ($R_5$).} Top-5 recall measures the overlap between the five highest predicted models and the five actual optimal models, 
{
\begin{equation}
    R_5 = \frac{|F_5|}{5}, \quad F_5 = I(\mathcal{A}^5)\cap I(\mathcal{S}^5),
\end{equation}
}
where $I(\cdot)$ is the index set, $\mathcal{A}^5$ and $\mathcal{S}^5$ are top-5 values in $\mathcal{A}$ and $\mathcal{S}$, $\|F_5\|$ is the length of $F_5$.
\item \textbf{Top-1 Accuracy.} Top-1 accuracy measures the reliability of a UVMS method when selecting a single model, which is defined as the ground truth accuracy of the model with the highest predicted score.
\item \textbf{Kendall's Rank Correlation ($\tau_5$ and $\tau$).} We use Kendall's rank correlation to evaluate the overall ranking ability of the UVMS method on the best five models and the entire model zoo,
\begin{equation}
\begin{split}
    \tau &= \frac{2}{|F|(|F|-1)} \sum_{i<j, \, i,j \in F} \operatorname{sign}(a_i-a_j) \operatorname{sign}(s_i-s_j), \\
    \tau_5 &= \left\{
    \begin{aligned}
        & 0,  \quad \text{if} \quad |F_5| < 2, \\
        &\frac{2}{|F_5|(|F_5|-1)} \sum_{\substack{i<j \\ i,j \in F_5}} \operatorname{sign}(a_i-a_j) \operatorname{sign}(s_i-s_j), \\
        &\quad \text{otherwise}.
    \end{aligned}
    \right.
\end{split}
\end{equation}
where $ F=\{1,2,...,M\}$, $F_5 = I(\mathcal{A}^5)\cap I(\mathcal{S}^5)$, and  $\operatorname{sign(\cdot)}$ is the sign function.
\end{itemize}

\section{Method}
\begin{figure*}[ht]
    \centering
    \includegraphics[width=1.\linewidth]{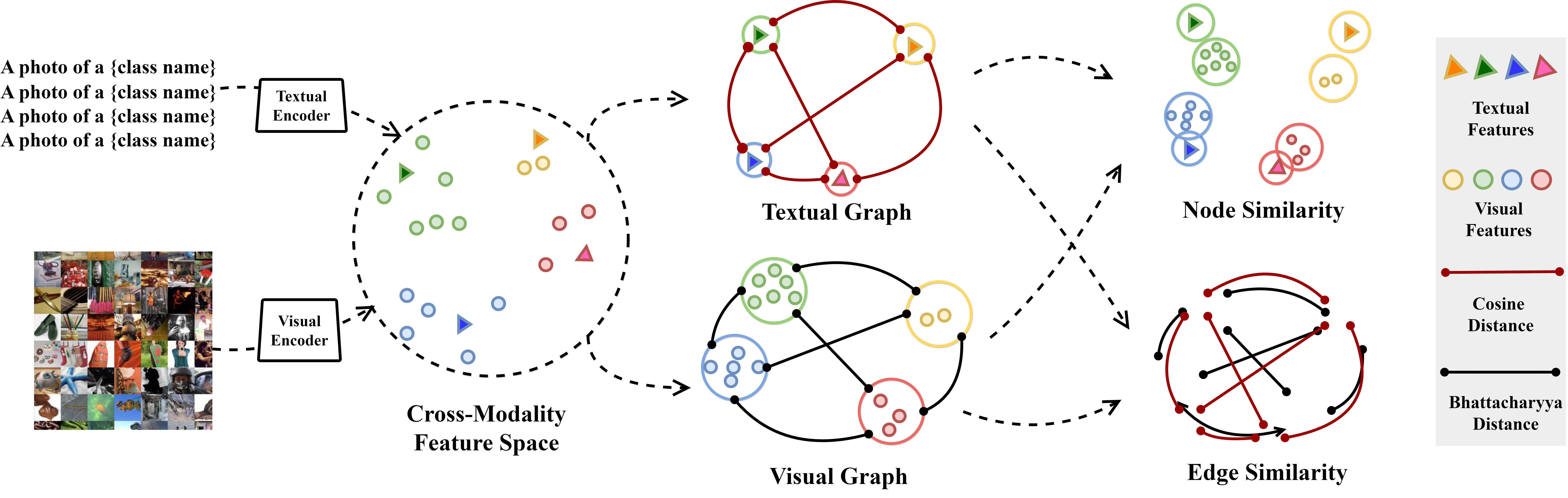}
    \caption{The pipeline of VEGA involves encoding class names and unlabeled images into a shared cross-modality feature space. Subsequently, we construct a textual graph and a visual graph for the two modalities, respectively. VEGA combines node-level and edge-level similarities to evaluate the alignment between these graphs.
}
    \label{fig:method}
\end{figure*}

\subsection{Motivation}
Vision language models have flourished in recent years \cite{radford2021learning,li2022blip,zhai2023sigmoid}.
The classical VLM pre-pretraining paradigm is based on contrastive learning techniques.
NT-Xent loss is extended to the multimodal domain,
\begin{equation}
\begin{aligned}
    \mathcal{L}_{\text{VLM}} = &-\frac{1}{2} \, \mathbb{E}_{(x, y) \sim P_{\text{data}},\{x'_i,y'_i\}^N_{i=1} \sim P_{\text{data}}} \\
&\left[ 
\log \frac{\exp(x^\top y / \tau)}{\sum_i \exp(x^{\prime\top}_i y / \tau)} 
+ 
\log \frac{\exp(x^\top y / \tau)}{\sum_i \exp(x^\top y'_i / \tau)} 
\right].    
\end{aligned}
\end{equation}
$\mathcal{L}_{VLM}$ aligns the positive pair of text $y$ and the corresponding image $x$.
Concurrently, $N$ negative pairs are denoted as $\{x'_i, y'_i\}$.
Both positive and negative pairs are sampled from the original data distribution $P_{\text{data}}$.
With contrastive pre-training, the features of both modalities are mapped into a shared representation space, where images and texts with the same semantics are clustered together.
Zero-shot classification is all about selecting the most recent class name for an image.
In the process of modal alignment, the performance of zero-shot classification is gradually improved.
Therefore, the performance of the VLM can be estimated by measuring the modality gap, \ie, the alignment level between modalities.

\subsection{VEGA: Visual-Textual Graph Alignment for Unsupervised VLM Selection}
The key idea behind our method is that in the shared cross-modality feature space, the more similar the structures of the class feature distributions are between the two modalities, the easier it becomes to match images with their corresponding classes.
Based on this intuition, we propose Visual-tExtual Graph Alignment (VEGA) to measure the similarity between these structures.
The pipeline of VEGA is shown in Fig. \ref{fig:paradigm}.
The class names are transformed into text prompts, which, along with unlabeled images, are encoded using the textual and visual encoders, respectively.
We then represent the structure of the class feature distributions for the two modalities as a textual graph and a vision graph.
VEGA is defined as the similarity between these two graphs.
The key challenge of VEGA is constructing modality-specific class distribution graphs and measuring their similarity. 
We will elaborate on these details in the following sections.

\noindent
\textbf{Textual Graph.}
Given the limited information available from the textual modality, we represent the nodes directly as the text features of each class and the edges as the cosine similarity between each pair of nodes.
Formally, the fully connected textual graph is denoted by $G_T=\{N_T,E_T\}$, where $N_T=\{n_k^T\}_{k=1}^K$ represents the nodes, and $E_T=\{e^T_{ij}\}_{i,j=1}^K$ represents the edges. 
Specifically, $n^T_k = \xi(\Tilde{c}_k)$ denotes the node features, and $e^T_{ij}=cos(\xi(\Tilde{c}_i), \xi(\Tilde{c}_j))$ denotes the edge weights, calculated as the cosine similarity between the textual features.

\noindent
\textbf{Visual Graph.}
Modeling the visual graph $G_V=\{N_V, E_V\}$, is more complex than modeling the the textual graph. 
Nodes cannot be represented by a single vector for two reasons.
First, a single vector lacks the capacity to fully represent a class. 
Second, without labels, it is challenging to determine which class an image belongs to.
Therefore, in the cross-modal feature space, we use $K$ textual features as centers to partition the visual features into $K$ clusters. 
The concatenation of features within each cluster represents a node: 
\begin{equation}
    n^V_k = cat(\{\phi(x_i)\cdot\mathbb{I}(\hat{y_i}=c_k )\}_{i=1}^N),
\end{equation}
where $\hat{y_i} = \mathop{argmax}\limits_k(cos(\xi_m(\Tilde{c_k}), \phi_m(x_i)))$, and $cat(\cdot)$ denotes concatenation.
Since the number of visual features in each class cluster varies, node sizes differ, making it unsuitable to use a simple cosine distance for calculating edges.
To address this issue, we model each class as a Gaussian distribution $\mathcal{N}_k$ with class means $\overline{n}^V_k$ and covariance $\Sigma_{k}$. $\overline{n}^V_k$ is the mean vector of $n^V_k$, and $\Sigma_{k}$ is the covariance matrix,
\begin{equation}
\begin{aligned}
    &\overline{n}^V_k = \frac{1}{N_k} \sum\limits_{i=1}^{N} \phi(x_i)\cdot\mathbb{I}(\hat{y_i}=c_k ), \\
    &\Sigma_k =\frac{1}{N_k} \sum_{i=1}^{N} \left(\phi(x_i)-\overline{n}^V_k\right)\left(\phi(x_i)-\overline{n}^V_k\right)^{\top} \cdot \mathbb{I}(\hat{y_i}=c_k ),
\end{aligned}
\end{equation}
where $N_k=\sum\limits_{i=1}^{N}\mathbb{I}(\hat{y_i}=c_k)$ is the number of features in the cluster of $c_k$.
Each edge $e^V_{ij}$ in $E_V=\{e^V_{ij}\}_{i,j=1}^K$ is defined by the Bhattacharyya coefficient between each pair of class Gaussians,
\begin{equation}
\begin{aligned}
    e^V_{ij} &= Bh(\mathcal{N}_i, \mathcal{N}_j) \\
        &=\frac{1}{8}\left(\overline{{n}}_{i}^V-\overline{n}_{j}^V\right)^{\top} \Sigma^{-1}\left(\overline{n}_{i}^V-\overline{n}_{j}^V\right) + \frac{1}{2} \ln \frac{|\Sigma|}{\sqrt{\left|\Sigma_{i}\right|\left|\Sigma_{j}\right|}},
\end{aligned}
\end{equation}
where $\Sigma = \frac{1}{2} \left (\Sigma_{i} + \Sigma_{j} \right)$, $\left|\cdot\right|$ denotes determinant.
Using distributional distance as the edge measure, rather than the distance between single vectors like class means, more accurately represents the relationships between classes. 
This approach accounts for within-class covariance, capturing the dispersion of features within each class.

\noindent
\textbf{Cross-Modality Graph Similarity.}
Finally, the VEGA score $s$ is defined as the summation of the node similarity $s_n$ and edge similarity $s_e$.
Node similarity is determined by the weighted average distance from all visual features within a cluster to the corresponding textual feature,
\begin{equation}
    s_n = \frac{1}{K} \sum\limits_{i=1}^K sim(n^T_k,n^V_k)\cdot N_k,
\end{equation}
where $N_k=\sum\limits_{i=1}^{N}\mathbb{I}(\hat{y_i}=c_k)$.
Considering the different scales of various VLM features, we normalize each feature at the class level to obtain relative distances,
\begin{equation}\label{eq:nodesim}
\begin{aligned}
    &sim(n^T_k,n^V_k) = \\
    &\frac{1}{N_k}\sum\limits_{i=1}^N \frac{exp \left( cos(\phi(x_i), \xi(c_k) ) / t \right)}{\sum\limits_{k'=1}^K exp \left( cos(\phi(x_i), \xi(c_{k'}) ) / t \right)}\cdot\mathbb{I}(\hat{y_i}=c_k ),
    \end{aligned}
\end{equation}
where $exp(\cdot)$ is the exponential function, and $t=0.05$ is a temperature parameter in the normalization.
For any VLM, the range of node similarity $s_n$ is constrained to the range of 0 to 1. 
Similarly, due to the scale differences between the Bhattacharyya coefficient and cosine distances, we use the Pearson correlation coefficient \cite{pearson1900mathematical} to measure edge similarity,
\begin{equation}
\begin{aligned}
    corr(E_T,E_V) &= \frac{\sum_{i=1}^{K^2} (e^V_i - \overline{e}^V)(e^T_i - \overline{e}^T)}{\sqrt{\sum_{i=1}^{K^2} (e^V_i - \overline{e}^V)^2} \sqrt{\sum_{i=1}^{K^2} (e^T_i - \overline{e}^T)^2}},
\end{aligned}
\end{equation}
where $e_i$ is $i$-th element in $E$, $\overline{e}^V$ and $\overline{e}^T$ denote the mean value of $E_V$ and $E_T$, respectively.
Since the Pearson correlation coefficient ranges from -1 to 1, we re-scale $s_e$ to a range of 0 to 1 to avoid the trade-off between $s_n$ and $s_e$, 
\begin{equation}
    s_e = \frac{1}{2}\cdot corr(E_T,E_V)+\frac{1}{2}.
\end{equation}
The formulation of VEGA is a simple summation of the two similarities: $s = s_n+s_e$.
VEGA is a user-friendly method, as its implementation requires no backward propagation process and does not rely on LLMs. 
It involves only a small amount of inference and computation, making it easy to implement on general mid-range to low-end GPUs and CPUs.

\section{Experiments}
We construct three benchmarks across three practical application scenarios for VLM performance evaluation, including performance prediction for VLMs from the CLIP family and various other pre-training algorithms respectively; and ranking the combinations of VLM and prompt templates.

\noindent
\textbf{Downstream Datasets.}
We conduct performance prediction on ten common-used downstream datasets, including basic image recognition Cifar-100 \cite{krizhevsky2009learning}; animal and plant dataset Oxford Pets \cite{parkhi2012cats} and Oxford Flowers \cite{nilsback2008automated}; street scene dataset SVHN \cite{netzer2011reading} and GTSRB \cite{houben2013detection}; describable textures dataset DTD \cite{cimpoi2014describing}; scene classification dataset Country211 \cite{thomee2016yfcc100m,radford2021learning} and SUN397 \cite{xiao2010sun}; digit dataset MNIST \cite{lecun1998gradient}; and facial expression dataset Fer2013 \cite{challenges-in-representation-learning-facial-expression-recognition-challenge}.

\noindent
\textbf{Baselines.}
We compare our method with existing training data-free methods in the fields of generalization error prediction \cite{hendrycks17baseline, deng2021does, xie2024importance}, unsupervised model validation \cite{saito2021tune}, and vision language model selection \cite{zohar2024lovm}.
These methods are highly related to our setting, which could evaluate the performance without training data and the annotations of downstream datasets.

\textbullet{} \textbf{Entropy (ENT)} is a commonly used baseline, representing the entropy of the probability distribution of the logits from VLMs,
\begin{equation}
\begin{aligned}
    &s_{ENT} = -\frac{1}{N}{\sum_{i=1}^N}P(x_i)\log P(x_i), \\
    &\text{where} \quad P(x_i) = \frac{exp \left( cos(\phi(x_i), \xi(c_k) ) \right)}{\sum\limits_{k'=1}^K exp \left( cos(\phi(x_i), \xi(c_{k'}) ) \right)}.    
\end{aligned}
\end{equation}

\textbullet{} \textbf{Confidence Score (Conf)} \cite{hendrycks17baseline} is a classical confidence-based method, defined as the average highest confidence score,
\begin{equation}
    s_{\text{Conf}} = \frac{1}{N}{\sum_{i=1}^N} max(\{P(x_i)[k]\}_{k=1}^{K}).
\end{equation}

\textbullet{} \textbf{Rotation (Rot)} \cite{deng2021does} is inspired by self-supervised methods \cite{gidaris2018unsupervised} and uses the accuracy of rotation angle prediction as a metric,
\begin{equation}
\begin{aligned}
    s_{\text{Rot}} &= \frac{1}{4N}{\sum_{i=1}^{4N}} \mathbb{I}(\hat{y}^r_i=y^r_i), \\ 
    \hat{y}_i^r &= \mathop{argmax}\limits_k(cos(\xi(Y^r_k), \phi(x_i))),
\end{aligned}
\end{equation}
where the label space $Y^r$ is defined as \{0, 90, 180, 270\}, and the template is `` An image rotated by {$y^r_i$} degrees''. 
Each image is augmented to obtain four rotated images.
Note that Rot can be used to select different image encoders, but not prompt templates, as image rotation does not involve text encoders.

\textbullet{} \textbf{SND} \cite{saito2021tune} is designed for unsupervised validation and is defined as neighborhood density,
\begin{equation}
\begin{aligned}
    s_{SND} &= -\frac{1}{N}{\sum_{i=1}^{N}} {\sum_{j=1}^{N}} D_{ij}\log D_{ij}, \\
    D_{ij} &= \frac{exp(N_{ij}/\tau)}{\sum_{j'}exp(N_{ij'}/\tau)},    
\end{aligned}
\end{equation}
where $N_{ij}$ is the cosine distance between i-th image and j-th image, \ie, soft neighbor distance.
SND measures the soft neighbor density of a representation space.

\textbullet{} \textbf{Dispersion Score (DS)} \cite{xie2024importance} performs unsupervised clustering on the target dataset, and measures the separability among class means,
\begin{equation}
    s_{DS} = \log \frac{\sum_{k=1}^K n_k \cdot \|\boldsymbol{\Bar{\mu}} - \widetilde{\boldsymbol{\mu}}_k\|_2^2}{K - 1},
\end{equation}
where $n_k$ is the number of k-th cluster, $\mu_k$ is the k-th cluster center, and $\Bar{\mu}$ is the center of cluster center.

\textbullet{} \textbf{LOVM-G} and \textbf{LOVM-C} \cite{zohar2024lovm} are the dataset granularity score and the text classification score respectively, which measure the dataset difficulty and class clarity.
LOVM-C leverages the generated captions dataset as image proxies and replace the images with the generated image captions to calculate each VLM’s text top-1 accuracy and f1-score.
LOVM-G includes three metrics: The Fisher criterion \cite{fisher1936use} evaluates the similarity and separation between classes; the Silhouette score \cite{rousseeuw1987silhouettes} quantifies the compactness of same-class samples relative to the separation of different-class samples; and Class Dispersion score, which is their normalization constant, measures the tightness within a single class or the radius of its data cone.
More details can be found in \cite{zohar2024lovm}.

\noindent
\textbf{Details.} All the experiments are conducted on NVIDIA Geforce 3090Ti GPU, and the temperature in Eq. (\ref{eq:nodesim}) set to 0.05 across all the experiments. 
There are no random operations involved in our experiments.
The Python implementation of VEGA and the specific predicted scores for all quantitative results are provided in the supplementary material.

\begin{table*}[ht]
\centering
\caption{Downstream zero-shot classification performance prediction on CLIP models with various architectures and source datasets.  \textcolor[HTML]{C00000}{Red} indicates the best result in each row, and \textcolor[HTML]{319B62}{green} represents the second-best. Oracle is the best accuracy in the candidate model, which is the upper bound of Top-1 Accuracy.}
\resizebox{1 \textwidth}{!}{ 
\setlength{\tabcolsep}{4pt}
\begin{tabular}{lccccccc>{\columncolor[HTML]{F3F5F7}}c|ccccccc>{\columncolor[HTML]{F3F5F7}}cc}
\cmidrule{1-17}
 & ENT & Conf & Rot & SND & DS & LOVM-G & LOVM-C & VEGA & ENT & Conf & Rot & SND & DS & LOVM-G & LOVM-C & VEGA &\\ \cmidrule{2-17} 
\multirow{-2}{*}{Dataset} & \multicolumn{8}{c|}{$R_5$} & \multicolumn{8}{c}{$\tau_5$} & \\ \cmidrule{1-17}
Cifar100  & \color[HTML]{319B62} 0.60 & \color[HTML]{319B62} 0.60 & 0.00 & 0.00 & 0.40 & 0.40 & 0.20 & \color[HTML]{C00000} 0.80 & -1.00 & -0.33 & \color[HTML]{319B62} 0.00 & \color[HTML]{319B62} 0.00 & \color[HTML]{C00000} 1.00 & \color[HTML]{C00000} 1.00 & \color[HTML]{319B62} 0.00 & \color[HTML]{C00000} 1.00 &\\
Country211  & \color[HTML]{319B62} 0.40 & \color[HTML]{C00000} 0.60 & \color[HTML]{319B62} 0.40 & 0.00 & 0.00 & 0.20 & 0.20 & \color[HTML]{C00000} 0.60 & -1.00 & -0.33 & \color[HTML]{C00000} 1.00 & 0.00 & 0.00 & 0.00 & 0.00 & \color[HTML]{319B62} 0.33 &\\
DTD  & \color[HTML]{319B62} 0.60 & \color[HTML]{C00000} 0.80 & 0.20 & 0.00 & 0.00 & 0.20 & \color[HTML]{319B62} 0.60 & \color[HTML]{C00000} 0.80 & -1.00 & -0.33 & 0.00 & 0.00 & 0.00 & 0.00 & \color[HTML]{C00000} 1.00 & \color[HTML]{319B62} 0.67 &\\
Flowers  & \color[HTML]{319B62} 0.60 & \color[HTML]{319B62} 0.60 & 0.40 & 0.00 & 0.00 & 0.00 & 0.00 & \color[HTML]{C00000} 0.80 & -0.33 & \color[HTML]{319B62} 0.33 & -1.00 & 0.00 & 0.00 & 0.00 & 0.00 & \color[HTML]{C00000} 0.67 &\\
GTSRB  & \color[HTML]{C00000} 0.60 & \color[HTML]{C00000} 0.60 & \color[HTML]{319B62} 0.20 & 0.00 & \color[HTML]{319B62} 0.20 & \color[HTML]{C00000} 0.60 & \color[HTML]{C00000} 0.60 & \color[HTML]{C00000} 0.60 & -0.33 & \color[HTML]{C00000} 0.33 & \color[HTML]{319B62} 0.00 & \color[HTML]{319B62} 0.00 & \color[HTML]{319B62} 0.00 & -0.33 & \color[HTML]{C00000} 0.33 & -0.33 &\\
MNIST  & 0.00 & \color[HTML]{319B62} 0.20 & \color[HTML]{C00000} 0.40 & 0.00 & \color[HTML]{319B62} 0.20 & 0.00 & \color[HTML]{319B62} 0.20 & \color[HTML]{C00000} 0.40 & \color[HTML]{319B62} 0.00 & \color[HTML]{319B62} 0.00 & -1.00 & \color[HTML]{319B62} 0.00 & \color[HTML]{319B62} 0.00 & \color[HTML]{319B62} 0.00 & \color[HTML]{319B62} 0.00 & \color[HTML]{C00000} 1.00 &\\
Pets  & \color[HTML]{319B62} 0.60 & \color[HTML]{319B62} 0.60 & 0.20 & 0.00 & 0.00 & 0.00 & 0.00 & \color[HTML]{C00000} 0.80 & \color[HTML]{319B62} 0.33 & -0.33 & 0.00 & 0.00 & 0.00 & 0.00 & 0.00 & \color[HTML]{C00000} 0.67 &\\
SVHN  & 0.20 & \color[HTML]{319B62} 0.40 & 0.20 & 0.00 & 0.20 & 0.20 & 0.00 & \color[HTML]{C00000} 0.80 & \color[HTML]{C00000} 0.00 & \color[HTML]{319B62} -1.00 & \color[HTML]{C00000} 0.00 & \color[HTML]{C00000} 0.00 & \color[HTML]{C00000} 0.00 & \color[HTML]{C00000} 0.00 & \color[HTML]{C00000} 0.00 & \color[HTML]{C00000} 0.00 &\\
SUN397  & \color[HTML]{319B62} 0.40 & \color[HTML]{C00000} 0.60 & 0.00 & 0.00 & 0.00 & 0.00 & \color[HTML]{319B62} 0.40 & \color[HTML]{319B62} 0.40 & \color[HTML]{C00000} 1.00 & \color[HTML]{C00000} 1.00 & \color[HTML]{319B62} 0.00 & \color[HTML]{319B62} 0.00 & \color[HTML]{319B62} 0.00 & \color[HTML]{319B62} 0.00 & \color[HTML]{C00000} 1.00 & -1.00 &\\
Fer2013  & \color[HTML]{C00000} 0.60 & \color[HTML]{C00000} 0.60 & 0.20 & 0.00 & \color[HTML]{C00000} 0.60 & \color[HTML]{C00000} 0.60 & 0.20 & \color[HTML]{319B62} 0.40 & \color[HTML]{319B62} -0.33 & \color[HTML]{319B62} -0.33 & \color[HTML]{C00000} 0.00 & \color[HTML]{C00000} 0.00 & -1.00 & -1.00 & \color[HTML]{C00000} 0.00 & -1.00 &\\ \cmidrule{1-17}
Avg. & 0.46 & \color[HTML]{319B62} 0.56 & 0.22 & 0.00 & 0.16 & 0.22 & 0.24 & \color[HTML]{C00000} 0.64 & -0.27 & -0.10 & -0.10 & 0.00 & 0.00 & -0.03 & \color[HTML]{C00000} 0.23 & \color[HTML]{319B62} 0.20 &\\
\cmidrule{1-17}
\\
\toprule
 & ENT & Conf & Rot & SND & DS & LOVM-G & LOVM-C & VEGA & ENT & Conf & Rot & SND & DS & LOVM-G & LOVM-C & VEGA & \\ \cmidrule{2-17} 
\multirow{-2}{*}{Dataset} & \multicolumn{8}{c|}{$\tau$} & \multicolumn{8}{c}{Top-1 Acc.} & Oracle \\ \midrule
Cifar100  & 0.51 & \color[HTML]{319B62} 0.63 & -0.09 & -0.54 & 0.54 & 0.09 & 0.34 & \color[HTML]{C00000} 0.81 & 0.78 & \color[HTML]{C00000} 0.85 & 0.72 & 0.40 & \color[HTML]{319B62} 0.80 & 0.78 & 0.78 & \color[HTML]{C00000} 0.85 & 0.85\\
Country211  & 0.48 & \color[HTML]{C00000} 0.59 & 0.17 & -0.32 & 0.16 & 0.19 & 0.51 & \color[HTML]{319B62} 0.57 & \color[HTML]{319B62} 0.29 & \color[HTML]{C00000} 0.30 & 0.21 & 0.15 & 0.22 & 0.22 & \color[HTML]{C00000} 0.30 & \color[HTML]{C00000} 0.30 & 0.33\\
DTD  & 0.57 & \color[HTML]{319B62} 0.69 & 0.11 & -0.43 & 0.45 & 0.15 & 0.53 & \color[HTML]{C00000} 0.77 & \color[HTML]{319B62} 0.66 & \color[HTML]{319B62} 0.66 & 0.53 & 0.35 & 0.59 & 0.58 & \color[HTML]{C00000} 0.68 & \color[HTML]{C00000} 0.68 & 0.68\\
Flowers  & 0.50 & \color[HTML]{319B62} 0.62 & 0.15 & -0.26 & 0.32 & 0.06 & 0.07 & \color[HTML]{C00000} 0.66 & \color[HTML]{319B62} 0.80 & \color[HTML]{319B62} 0.80 & 0.73 & 0.55 & 0.72 & 0.73 & 0.58 & \color[HTML]{C00000} 0.81 & 0.81\\
GTSRB  & 0.36 & \color[HTML]{C00000} 0.48 & 0.09 & -0.17 & 0.34 & 0.44 & \color[HTML]{319B62} 0.47 & 0.46 & 0.47 & \color[HTML]{C00000} 0.50 & 0.47 & 0.29 & \color[HTML]{319B62} 0.49 & 0.44 & 0.47 & \color[HTML]{C00000} 0.50 & 0.55\\
MNIST  & 0.26 & 0.34 & 0.07 & -0.11 & 0.20 & 0.22 & \color[HTML]{319B62} 0.46 & \color[HTML]{C00000} 0.50 & 0.56 & 0.56 & 0.29 & 0.66 & \color[HTML]{319B62} 0.69 & \color[HTML]{319B62} 0.69 & \color[HTML]{C00000} 0.73 & 0.58 & 0.76\\
Pets  & 0.56 & \color[HTML]{319B62} 0.64 & 0.25 & -0.44 & 0.37 & 0.04 & 0.05 & \color[HTML]{C00000} 0.74 & \color[HTML]{319B62} 0.92 & \color[HTML]{319B62} 0.92 & 0.91 & 0.75 & 0.88 & 0.88 & 0.91 & \color[HTML]{C00000} 0.93 & 0.93\\
SVHN  & \color[HTML]{319B62} 0.47 & 0.44 & -0.19 & -0.33 & 0.38 & 0.38 & -0.05 & \color[HTML]{C00000} 0.56 & \color[HTML]{319B62} 0.46 & \color[HTML]{319B62} 0.46 & 0.34 & 0.16 & \color[HTML]{319B62} 0.46 & \color[HTML]{C00000} 0.50 & 0.45 & \color[HTML]{319B62} 0.46 & 0.56\\
SUN397  & 0.62 & \color[HTML]{319B62} 0.72 & -0.34 & -0.49 & 0.30 & 0.10 & 0.41 & \color[HTML]{C00000} 0.78 & \color[HTML]{C00000} 0.76 & \color[HTML]{C00000} 0.76 & 0.64 & 0.50 & 0.69 & \color[HTML]{319B62} 0.72 & \color[HTML]{319B62} 0.72 & \color[HTML]{C00000} 0.76 & 0.76\\
Fer2013  & 0.32 & \color[HTML]{319B62} 0.37 & 0.10 & -0.35 & 0.28 & 0.04 & \color[HTML]{C00000} 0.44 & 0.33 & 0.28 & \color[HTML]{319B62} 0.33 & 0.32 & 0.23 & 0.28 & \color[HTML]{319B62} 0.33 & 0.32 & \color[HTML]{C00000} 0.34 & 0.34\\ \midrule
Avg. & 0.46 & \color[HTML]{319B62} 0.55 & 0.03 & -0.35 & 0.33 & 0.17 & 0.32 & \color[HTML]{C00000} 0.62 & \color[HTML]{319B62} 0.60 & \color[HTML]{C00000} 0.62 & 0.52 & 0.40 & 0.58 & 0.59 & 0.59 & \color[HTML]{C00000} 0.62 & 0.66\\
\midrule
\end{tabular}
}
\label{tab:arch}
\end{table*}

\subsection{Performance Prediction for VLMs from CLIP Family}\label{sec:main}
CLIP \cite{radford2021learning} is the most popular VLM in recent years, with many CLIP models trained on various architectures and source datasets available as open-source.
We first examine the evaluation of the CLIP family across different network architectures and source datasets.

\noindent
\textbf{Candidate Models.}
We have collected 31 CLIP models with diverse architectures and source datasets from OpenCLIP \footnote{https://github.com/mlfoundations/open\_clip}.
These models are the same as those used in LOVM \cite{zohar2024lovm} and include architectures from model families such as ResNet \cite{he2016deep}, ViT \cite{dosovitskiy2021an}, and ConvNeXt \cite{liu2022convnet}, with source datasets comprising various versions of LAION \cite{schuhmann2021laion}. 
Detailed information on the candidate models is provided in the supplementary material. 
For all candidate models, we use several commonly used prompt templates \cite{zohar2024lovm} and compute the mean to obtain the textual feature.

\noindent
\textbf{Quantitative Results.}
We provide the complete results in Table \ref{tab:arch}, where red indicates the best result in each row, and green represents the second-best.
The table showcases the effectiveness of different methods in predicting downstream performance across various datasets on the CLIP family. 
Specifically, VEGA achieves the highest average scores for both Top-5 recall ($R_5$) and overall Kendall correlation (\(\tau\)), with values of 0.64 and 0.62, respectively, showcasing its robustness in model selection. 
Notably, VEGA consistently ranks first or second in most datasets, including Flowers, GTSRB, and OxfordPets, where accurate predictions are critical for selecting the best-performing models. 
Additionally, VEGA's Top-1 accuracy aligns closely with the Oracle results, further validating its reliability in identifying models with superior downstream performance. 
These results highlight VEGA as a state-of-the-art, user-friendly approach for VLM selection and performance prediction.
SND shows a negative correlation with downstream tasks, which might be because SND is designed for unsupervised validation tasks.
However, in VLM model selection, where the differences between models are often more pronounced, a higher SND might indicate that the model has not learned a clear decision boundary.

\begin{figure*}[ht]
    \centering
    \includegraphics[width=.7\linewidth]{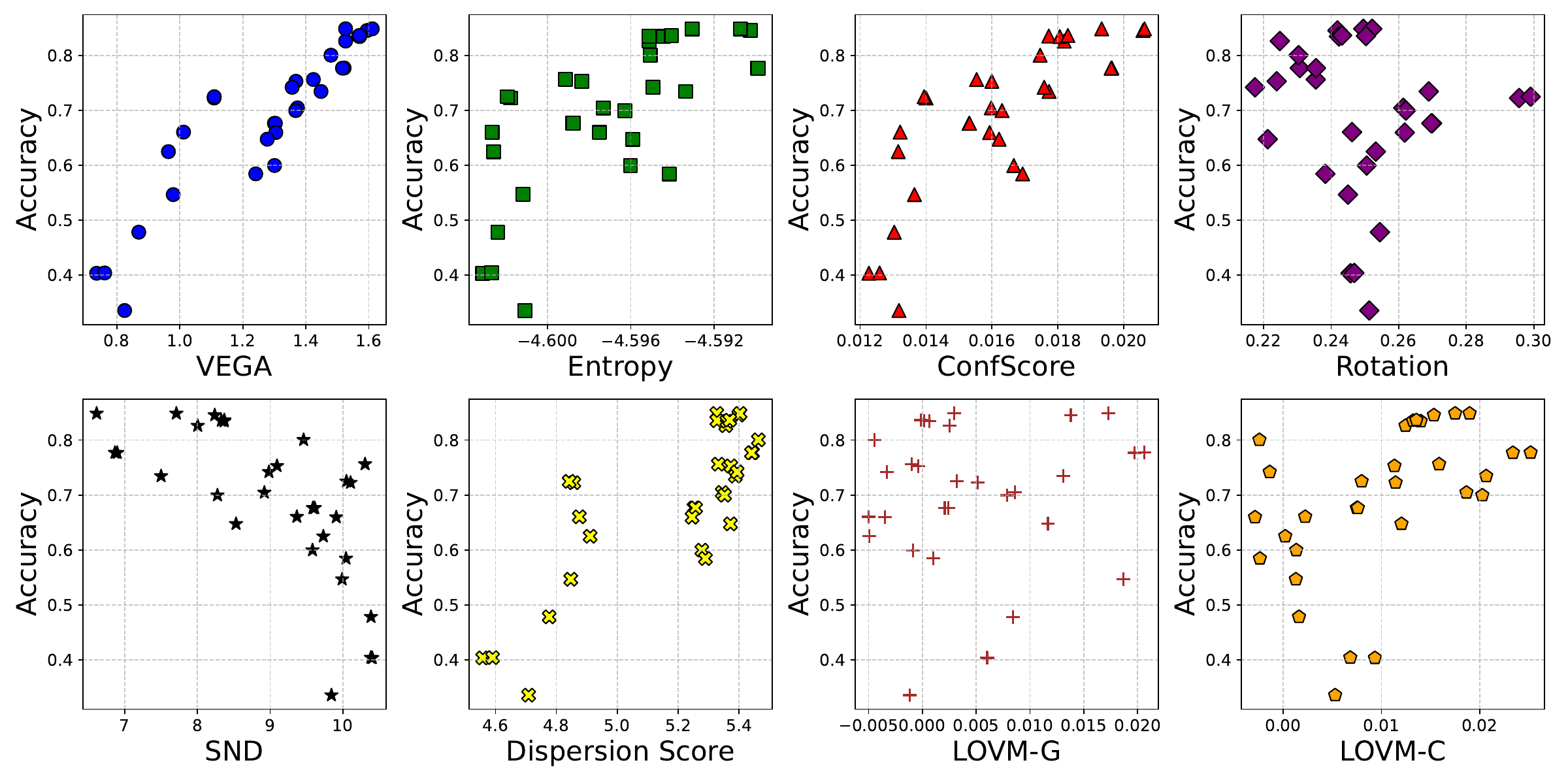}
    \caption{Visualization of the correlation between the actual zero-shot classification accuracy and predicted scores for various VLMs in the CLIP family.}
    \label{fig:main}
\end{figure*}

\noindent
\textbf{Qualitative Results.}
We visualize the correlation between the actual downstream accuracy and the predicted scores for various methods in Fig. \ref{fig:main}. 
The overall trend is basically consistent with that of quantitative experiments.
Our method shows the strongest correlation, with data points closely following a linear trend, indicating high predictive accuracy. 
In contrast, methods like Entropy, Confidence Score, and Dispersion Score exhibit moderate correlations with more scattered data points, reflecting less consistent predictive power. 
Rotation and SND display the weakest correlations, with widely dispersed points and no clear linear pattern. 

\begin{table*}[ht]
\centering
\caption{Downstream zero-shot classification performance prediction for VLMs from various pre-training algorithms.}
\resizebox{1. \textwidth}{!}{ 
\setlength{\tabcolsep}{4pt}
\begin{tabular}{lccccc>{\columncolor[HTML]{F3F5F7}}c|ccccc>{\columncolor[HTML]{F3F5F7}}c|ccccc>{\columncolor[HTML]{F3F5F7}}c|ccccc>{\columncolor[HTML]{F3F5F7}}cc}
\toprule
& ENT & Conf & Rot & SND & DS & VEGA & ENT & Conf & Rot & SND & DS & VEGA & ENT & Conf & Rot & SND & DS & VEGA & ENT & Conf & Rot & SND & DS & VEGA &  \\ \cmidrule{2-26} 
\multirow{-2}{*}{Dataset} & \multicolumn{6}{c|}{$R_5$} & \multicolumn{6}{c|}{$\tau_5$}  & \multicolumn{6}{c|}{$\tau$} & \multicolumn{6}{c}{Top-1 Acc.} & Oracle\\ \midrule
Cifar100  & \color[HTML]{319B62} 0.40 & \color[HTML]{319B62} 0.40 & \color[HTML]{319B62} 0.40 & 0.20 & \color[HTML]{319B62} 0.40 & \color[HTML]{C00000} 0.60 & -1.00 & -1.00 & -1.00 & \color[HTML]{319B62} 0.00 & -1.00 & \color[HTML]{C00000} 1.00 & 0.19 & \color[HTML]{319B62} 0.40 & 0.03 & -0.21 & 0.35 & \color[HTML]{C00000} 0.71 & \color[HTML]{319B62} 0.80 & \color[HTML]{319B62} 0.80 & 0.72 & 0.09 & \color[HTML]{319B62} 0.80 & \color[HTML]{C00000} 0.82 & 0.82\\
Country211  & \color[HTML]{319B62} 0.60 & \color[HTML]{C00000} 0.80 & \color[HTML]{319B62} 0.60 & 0.20 & 0.00 & \color[HTML]{C00000} 0.80 & -1.00 & -0.33 & \color[HTML]{C00000} 0.33 & \color[HTML]{319B62} 0.00 & \color[HTML]{319B62} 0.00 & -0.67 & 0.41 & \color[HTML]{319B62} 0.44 & 0.25 & -0.22 & 0.24 & \color[HTML]{C00000} 0.51 & \color[HTML]{319B62} 0.29 & \color[HTML]{319B62} 0.29 & \color[HTML]{C00000} 0.32 & 0.03 & 0.01 & \color[HTML]{319B62} 0.29 & 0.33\\
DTD  & 0.20 & 0.20 & \color[HTML]{C00000} 0.60 & \color[HTML]{319B62} 0.40 & 0.20 & \color[HTML]{319B62} 0.40 & \color[HTML]{C00000} 1.00 & \color[HTML]{319B62} 0.67 & 0.00 & 0.00 & \color[HTML]{319B62} 0.67 & \color[HTML]{319B62} 0.67 & 0.59 & \color[HTML]{319B62} 0.68 & -0.06 & -0.24 & 0.56 & \color[HTML]{C00000} 0.72 & \color[HTML]{C00000} 0.68 & \color[HTML]{C00000} 0.68 & \color[HTML]{319B62} 0.51 & 0.10 & \color[HTML]{C00000} 0.68 & \color[HTML]{C00000} 0.68 & 0.68\\
Flowers  & 0.20 & 0.20 & \color[HTML]{C00000} 0.60 & \color[HTML]{319B62} 0.40 & 0.20 & \color[HTML]{319B62} 0.40 & \color[HTML]{C00000} 1.00 & \color[HTML]{C00000} 1.00 & -1.00 & -1.00 & \color[HTML]{319B62} 0.00 & \color[HTML]{C00000} 1.00 & 0.47 & \color[HTML]{319B62} 0.50 & 0.10 & -0.25 & 0.26 & \color[HTML]{C00000} 0.68 & 0.69 & 0.69 & \color[HTML]{319B62} 0.73 & 0.07 & 0.55 & \color[HTML]{C00000} 0.88 & 0.88\\
GTSRB  & 0.20 & 0.20 & \color[HTML]{C00000} 0.60 & \color[HTML]{319B62} 0.40 & \color[HTML]{319B62} 0.40 & \color[HTML]{C00000} 0.60 & -0.33 & -0.33 & \color[HTML]{C00000} 1.00 & 0.00 & 0.33 & \color[HTML]{319B62} 0.60 & 0.43 & \color[HTML]{319B62} 0.49 & 0.18 & 0.06 & 0.46 & \color[HTML]{C00000} 0.65 & \color[HTML]{319B62} 0.48 & \color[HTML]{319B62} 0.48 & 0.40 & 0.07 & \color[HTML]{319B62} 0.48 & \color[HTML]{C00000} 0.64 & 0.64\\
MNIST  & \color[HTML]{C00000} 0.80 & \color[HTML]{C00000} 0.80 & \color[HTML]{319B62} 0.40 & 0.00 & \color[HTML]{C00000} 0.80 & \color[HTML]{C00000} 0.80 & 0.00 & \color[HTML]{C00000} 1.00 & \color[HTML]{C00000} 1.00 & 0.00 & \color[HTML]{319B62} 0.33 & \color[HTML]{C00000} 1.00 & \color[HTML]{319B62} 0.59 & \color[HTML]{C00000} 0.68 & 0.16 & -0.53 & \color[HTML]{319B62} 0.59 & \color[HTML]{319B62} 0.59 & 0.77 & \color[HTML]{C00000} 0.88 & 0.65 & 0.08 & \color[HTML]{319B62} 0.81 & \color[HTML]{C00000} 0.88 & 0.88\\
Pets  & 0.40 & \color[HTML]{319B62} 0.60 & 0.40 & 0.20 & 0.40 & \color[HTML]{C00000} 0.80 & -0.33 & 0.33 & \color[HTML]{C00000} 1.00 & 0.00 & -0.33 & \color[HTML]{319B62} 0.67 & 0.51 & \color[HTML]{319B62} 0.61 & 0.10 & -0.49 & 0.52 & \color[HTML]{C00000} 0.82 & \color[HTML]{319B62} 0.91 & \color[HTML]{319B62} 0.91 & 0.90 & 0.03 & \color[HTML]{319B62} 0.91 & \color[HTML]{C00000} 0.95 & 0.95\\
SVHN  & \color[HTML]{319B62} 0.60 & \color[HTML]{319B62} 0.60 & \color[HTML]{319B62} 0.60 & 0.00 & 0.20 & \color[HTML]{C00000} 0.80 & 0.00 & -0.33 & -1.00 & 0.00 & \color[HTML]{C00000} 1.00 & \color[HTML]{319B62} 0.67 & \color[HTML]{319B62} 0.59 & 0.53 & 0.18 & -0.34 & 0.43 & \color[HTML]{C00000} 0.66 & \color[HTML]{C00000} 0.47 & \color[HTML]{C00000} 0.47 & \color[HTML]{319B62} 0.42 & 0.07 & \color[HTML]{C00000} 0.47 & \color[HTML]{C00000} 0.47 & 0.48\\
SUN397  & \color[HTML]{319B62} 0.40 & \color[HTML]{C00000} 0.60 & \color[HTML]{319B62} 0.40 & 0.20 & \color[HTML]{319B62} 0.40 & \color[HTML]{C00000} 0.60 & \color[HTML]{C00000} 0.67 & \color[HTML]{319B62} 0.60 & 0.00 & 0.00 & 0.33 & \color[HTML]{C00000} 0.67 & 0.38 & \color[HTML]{319B62} 0.65 & -0.03 & -0.31 & 0.43 & \color[HTML]{C00000} 0.82 & 0.04 & \color[HTML]{C00000} 0.75 & \color[HTML]{319B62} 0.61 & 0.19 & 0.53 & \color[HTML]{C00000} 0.75 & 0.75\\
Fer2013  & 0.20 & 0.20 & \color[HTML]{C00000} 0.60 & 0.20 & \color[HTML]{319B62} 0.40 & \color[HTML]{C00000} 0.60 & \color[HTML]{C00000} 1.00 & \color[HTML]{C00000} 1.00 & -1.00 & \color[HTML]{319B62} 0.00 & \color[HTML]{C00000} 1.00 & -1.00 & 0.24 & 0.21 & 0.07 & -0.47 & \color[HTML]{C00000} 0.44 & \color[HTML]{319B62} 0.28 & 0.28 & \color[HTML]{C00000} 0.36 & 0.30 & 0.26 & 0.28 & \color[HTML]{319B62} 0.31 & 0.36\\
\midrule
Avg. & 0.40 & 0.46 & \color[HTML]{319B62} 0.52 & 0.22 & 0.34 & \color[HTML]{C00000} 0.64 & 0.10 & \color[HTML]{319B62} 0.26 & -0.07 & -0.10 & 0.23 & \color[HTML]{C00000} 0.46 & 0.44 & \color[HTML]{319B62} 0.52 & 0.10 & -0.30 & 0.43 & \color[HTML]{C00000} 0.64 & 0.54 & \color[HTML]{319B62} 0.63 & 0.56 & 0.10 & 0.55 & \color[HTML]{C00000} 0.67 & 0.67\\
\bottomrule
\end{tabular}
}
\label{tab:mm}
\end{table*}

\subsection{Performance Prediction for VLMs from Various Pre-training Algorithms}\label{sec:mm}
Recent advancements in VLM algorithms have created a range of options for users.
When selecting an algorithm for zero-shot classification, which typically involves a standard network structure (visual and textual encoders), performance prediction methods can be applied similarly to those used for CLIP.
LOVM-G and LOVM-C are not compared because they did not provide the caption and synonym datasets generated by LLMs. 
In our experiments, we found that the effects of LOVM-G and LOVM-C are sensitive to the caption and synonym datasets, so we cannot guarantee the reliability of our reproducible results.

\noindent
\textbf{Candidate Models.} We collect 17 models from Hugging Face \footnote{https://huggingface.co} from 10 commonly used VLM pre-training algorithms on zero-shot classification, including ALIGN \cite{jia2021scaling}, AltCLIP \cite{chen-etal-2023-altclip}, CLIP \cite{radford2021learning}, GroupViT \cite{xu2022groupvit}, SigLIP \cite{zhai2023sigmoid}, StreetCLIP \cite{haas2023learning}, MetaClIP\cite{xu2023demystifying}, BiomedCLIP \cite{biomedclip}, QuiltNet \cite{ikezogwo2023quilt1m}, BioCLIP \cite{stevens2024bioclip}. 
For each algorithm, we select two models, except for those methods that only have one official open-source model.
In total, there are 17 models, with specific information provided in the supplementary material.

\noindent
\textbf{Quantitative Results.}
We compare the performance of various methods in predicting downstream performance based on the pre-training algorithms of VLMs in Table \ref{tab:mm}. 
VEGA achieves the highest average performance across four metrics.
The baseline method Confidence Score also performs well on several simple datasets, likely because the dataset has smaller inter-class differences, leading to higher model uncertainty.
In contrast, other methods exhibit weaker and less consistent performance. 
A high \( R_5 \) score (0.52) for Rotation, combined with average performance on other metrics, indicates that rotation is relatively reliable when selecting a few high-performing models.
SND still shows a negative average correlation ($\tau$ of -0.30), indicating poor alignment with actual downstream results. 
DS and ENT perform better than SND, but they are not as effective as Rot and VEGA.
Overall, VEGA’s strong and consistent performance across various datasets underscores its effectiveness in predicting the downstream impact of VLM pre-training algorithms, making it a more reliable and accurate method compared to its counterparts.

\begin{figure*}
    \centering
    \includegraphics[width=1.\linewidth]{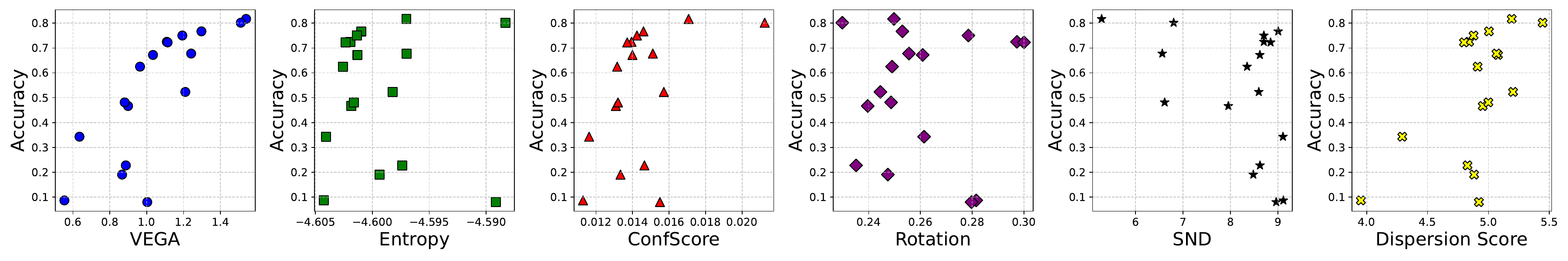}
    \caption{Visualization of correlations between the actual zero-shot classification accuracy and the predicted scores for VLMs from various popular pre-training algorithms.}
    \label{fig:mm}
\end{figure*}

\noindent
\textbf{Qualitative Results.}
The visualization results are shown in Fig. \ref{fig:mm}.
VEGA exhibits a clear linear trend, and the DS points are also distributed along the diagonal. 
In contrast, the linear trends for other methods are less pronounced, especially Entropy and Confidence Score perform poorly in this setting.

\begin{figure*}[ht]
    \centering
    \includegraphics[width=1\linewidth]{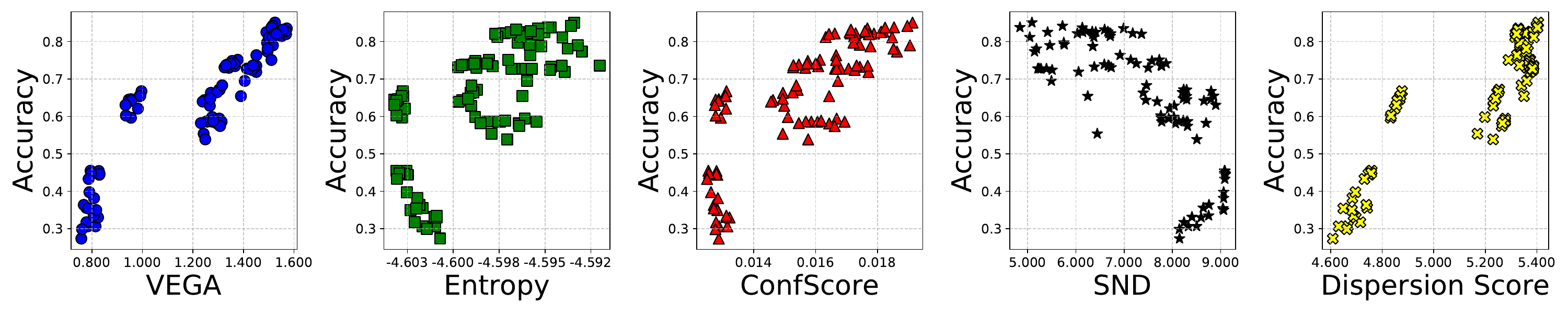}
    \caption{Visualization of correlations between the actual downstream accuracy and the predicted scores across combinations of model and prompt template.}
    \label{fig:mix}
\end{figure*}

\subsection{Performance Prediction for Combinations of VLM and Prompt Template}\label{sec:mix}
In practical VLM usage, selecting both a suitable model and an appropriate prompt template is essential. 
Thus we conduct experiments to evaluate the performance of different combinations of templates and models.
Note that Rotation \cite{deng2021does} is not included in this comparison, as its calculation pertains only to the image encoder and does not account for variations in prompt templates.
LOVM-G and LOVM-C are also as we explained in Sec. \ref{sec:mm}.

\noindent
\textbf{Candidate Combinations.}
We select 10 CLIP models from open clip, including diverse network architectures and source datasets.
The prompt templates are generated by GPT \cite{openai2023gpt4}, with 10 different templates from simple to complex, short to long.
There are a total of 10$\times$10=100 combinations of model and template.
Detailed information is provided in the supplementary material.

\begin{table*}[ht]
\centering
\caption{Downstream zero-shot classification performance prediction for combinations of CLIP models and prompt templates.}
\resizebox{1 \textwidth}{!}{ 
\setlength{\tabcolsep}{4pt}
\begin{tabular}{lcccc>{\columncolor[HTML]{F3F5F7}}c|cccc>{\columncolor[HTML]{F3F5F7}}c|cccc>{\columncolor[HTML]{F3F5F7}}c|cccc>{\columncolor[HTML]{F3F5F7}}cc}
\toprule
& ENT & Conf & SND & DS & VEGA & ENT & Conf & SND & DS & VEGA & ENT & Conf & SND & DS & VEGA & ENT & Conf & SND & DS & VEGA &  \\ \cmidrule{2-22} 
\multirow{-2}{*}{Dataset} & \multicolumn{5}{c|}{$R_5$} & \multicolumn{5}{c|}{$\tau_5$}  & \multicolumn{5}{c|}{$\tau$} & \multicolumn{5}{c}{Top-1 Acc.} & Oracle\\ \midrule
Cifar100  & 0.40 & \color[HTML]{319B62} 0.60 & 0.00 & \color[HTML]{C00000} 0.80 & 0.20 & \color[HTML]{C00000} 1.00 & \color[HTML]{C00000} 1.00 & \color[HTML]{319B62} 0.00 & \color[HTML]{319B62} 0.00 & \color[HTML]{319B62} 0.00 & 0.46 & \color[HTML]{319B62} 0.61 & -0.56 & \color[HTML]{319B62} 0.61 & \color[HTML]{C00000} 0.77 & 0.74 & \color[HTML]{C00000} 0.85 & 0.45 & \color[HTML]{C00000} 0.85 & \color[HTML]{319B62} 0.84 & 0.85\\
Country211  & 0.40 & \color[HTML]{319B62} 0.80 & 0.00 & 0.00 & \color[HTML]{C00000} 1.00 & \color[HTML]{C00000} 1.00 & 0.00 & 0.00 & 0.00 & \color[HTML]{319B62} 0.60 & 0.33 & \color[HTML]{319B62} 0.46 & -0.35 & 0.30 & \color[HTML]{C00000} 0.47 & \color[HTML]{319B62} 0.22 & \color[HTML]{C00000} 0.28 & 0.17 & 0.18 & \color[HTML]{C00000} 0.28 & 0.28\\
DTD  & \color[HTML]{319B62} 0.40 & \color[HTML]{C00000} 0.60 & 0.00 & 0.00 & \color[HTML]{C00000} 0.60 & \color[HTML]{C00000} 1.00 & \color[HTML]{C00000} 1.00 & \color[HTML]{319B62} 0.00 & \color[HTML]{319B62} 0.00 & -0.33 & 0.52 & \color[HTML]{319B62} 0.60 & -0.34 & 0.55 & \color[HTML]{C00000} 0.73 & \color[HTML]{C00000} 0.65 & \color[HTML]{C00000} 0.65 & 0.44 & 0.57 & \color[HTML]{319B62} 0.64 & 0.65\\
Flowers  & \color[HTML]{319B62} 0.40 & \color[HTML]{C00000} 0.60 & 0.00 & 0.20 & \color[HTML]{C00000} 0.60 & \color[HTML]{C00000} 1.00 & -0.82 & \color[HTML]{319B62} 0.00 & -1.00 & -0.33 & 0.49 & 0.56 & -0.45 & \color[HTML]{319B62} 0.60 & \color[HTML]{C00000} 0.61 & \color[HTML]{C00000} 0.80 & 0.78 & 0.54 & 0.69 & \color[HTML]{319B62} 0.79 & 0.80\\
GTSRB  & \color[HTML]{319B62} 0.00 & \color[HTML]{C00000} 0.20 & \color[HTML]{319B62} 0.00 & \color[HTML]{319B62} 0.00 & \color[HTML]{C00000} 0.20 & \color[HTML]{C00000} 0.00 & \color[HTML]{C00000} 0.00 & \color[HTML]{C00000} 0.00 & \color[HTML]{C00000} 0.00 & \color[HTML]{C00000} 0.00 & 0.41 & \color[HTML]{319B62} 0.55 & -0.30 & 0.45 & \color[HTML]{C00000} 0.56 & \color[HTML]{319B62} 0.43 & \color[HTML]{C00000} 0.48 & 0.29 & 0.38 & \color[HTML]{C00000} 0.48 & 0.51\\
MNIST  & \color[HTML]{C00000} 0.00 & \color[HTML]{C00000} 0.00 & \color[HTML]{C00000} 0.00 & \color[HTML]{C00000} 0.00 & \color[HTML]{C00000} 0.00 & \color[HTML]{C00000} 0.00 & \color[HTML]{C00000} 0.00 & \color[HTML]{C00000} 0.00 & \color[HTML]{C00000} 0.00 & \color[HTML]{C00000} 0.00 & 0.23 & 0.33 & -0.20 & \color[HTML]{C00000} 0.42 & \color[HTML]{319B62} 0.37 & \color[HTML]{319B62} 0.46 & 0.37 & 0.41 & \color[HTML]{C00000} 0.61 & \color[HTML]{319B62} 0.46 & 0.71\\
Pets  & \color[HTML]{C00000} 0.80 & \color[HTML]{C00000} 0.80 & \color[HTML]{319B62} 0.00 & \color[HTML]{319B62} 0.00 & \color[HTML]{C00000} 0.80 & \color[HTML]{C00000} 0.00 & \color[HTML]{319B62} -0.40 & \color[HTML]{C00000} 0.00 & \color[HTML]{C00000} 0.00 & -0.60 & 0.60 & \color[HTML]{319B62} 0.64 & -0.49 & 0.61 & \color[HTML]{C00000} 0.73 & \color[HTML]{C00000} 0.91 & \color[HTML]{C00000} 0.91 & 0.83 & \color[HTML]{319B62} 0.86 & \color[HTML]{C00000} 0.91 & 0.91\\
SVHN  & \color[HTML]{319B62} 0.00 & \color[HTML]{319B62} 0.00 & \color[HTML]{319B62} 0.00 & \color[HTML]{319B62} 0.00 & \color[HTML]{C00000} 0.20 & \color[HTML]{C00000} 0.00 & \color[HTML]{C00000} 0.00 & \color[HTML]{C00000} 0.00 & \color[HTML]{C00000} 0.00 & \color[HTML]{C00000} 0.00 & 0.42 & \color[HTML]{319B62} 0.48 & -0.39 & 0.40 & \color[HTML]{C00000} 0.55 & \color[HTML]{319B62} 0.36 & \color[HTML]{C00000} 0.38 & 0.21 & 0.35 & \color[HTML]{C00000} 0.38 & 0.49\\
SUN397  & \color[HTML]{C00000} 0.00 & \color[HTML]{C00000} 0.00 & \color[HTML]{C00000} 0.00 & \color[HTML]{C00000} 0.00 & \color[HTML]{C00000} 0.00 & \color[HTML]{C00000} 0.00 & \color[HTML]{C00000} 0.00 & \color[HTML]{C00000} 0.00 & \color[HTML]{C00000} 0.00 & \color[HTML]{C00000} 0.00 & 0.46 & \color[HTML]{319B62} 0.60 & -0.37 & 0.44 & \color[HTML]{C00000} 0.69 & 0.68 & 0.68 & 0.52 & \color[HTML]{319B62} 0.70 & \color[HTML]{C00000} 0.72 & 0.74\\
Fer2013  & \color[HTML]{319B62} 0.00 & \color[HTML]{319B62} 0.00 & \color[HTML]{C00000} 0.40 & \color[HTML]{319B62} 0.00 & \color[HTML]{319B62} 0.00 & \color[HTML]{C00000} 0.00 & \color[HTML]{C00000} 0.00 & \color[HTML]{319B62} -1.00 & \color[HTML]{C00000} 0.00 & \color[HTML]{C00000} 0.00 & -0.05 & 0.02 & -0.07 & \color[HTML]{319B62} 0.05 & \color[HTML]{C00000} 0.11 & 0.18 & 0.18 & \color[HTML]{319B62} 0.31 & 0.21 & \color[HTML]{C00000} 0.32 & 0.35\\
\midrule
Avg. & \color[HTML]{319B62} 0.24 & \color[HTML]{C00000} 0.36 & 0.04 & 0.10 & \color[HTML]{C00000} 0.36 & \color[HTML]{C00000} 0.40 & \color[HTML]{319B62} 0.08 & -0.10 & -0.10 & -0.07 & 0.39 & \color[HTML]{319B62} 0.48 & -0.35 & 0.44 & \color[HTML]{C00000} 0.56 & 0.54 & \color[HTML]{319B62} 0.55 & 0.42 & 0.54 & \color[HTML]{C00000} 0.58 & 0.63\\
\bottomrule
\end{tabular}
}
\label{tab:mix}
\end{table*}

\noindent
\textbf{Quantitative Results.}
We compare the performance of different methods in predicting downstream results across various CLIP model and prompt template combinations, as shown in Table \ref{tab:mix} . 
VEGA achieves the highest average $R_5$ of 0.36, $\tau$ of 0.56 and Top-1 accuracy of 0.58, demonstrating superior predictive accuracy across all downstream datasets.
There are a lot of 0 results in $\tau_5$, which is because the task is difficult, and the Top-5 recall ($R_5$) is low overall, so $\tau_5$ is also low. 
Entropy performs well on this metric.
Confidence Score also performs well overall, being the best on $R_5$ and second-best on the remaining three metrics.
VEGA's consistent top performance across diverse datasets underscores its effectiveness in accurately predicting downstream performance for CLIP models and prompt template combinations.

\noindent
\textbf{Qualitative Results.}
Visualization of the correlations between actual downstream accuracy and predicted scores for comparison methods is shown in Fig. \ref{fig:mix}. 
The scatter plot for VEGA demonstrates a strong, positive linear correlation, with data points closely aligning along a diagonal line, indicating high predictive accuracy. 
In contrast, the plots for Entropy and SND show scattered patterns with no clear linear trend, reflecting weak correlations and lower predictive reliability. 
Confidence Score and Dispersion Score exhibit moderate correlations with some linearity, but their data points are more dispersed compared to VEGA. 

\begin{table*}[ht]
\centering
\caption{Ablation Study of VEGA on three benchmarks mentioned above: Performance Prediction for (a) VLMs from CLIP family (Sec. \ref{sec:main}); (b) VLMs from various pre-training algorithms (Sec. \ref{sec:mm}); and (c) combinations of VLM and prompt template (Sec. \ref{sec:mix}). Each row represents the average results on these benchmarks.}
\resizebox{0.8 \textwidth}{!}{ 
\setlength{\tabcolsep}{8pt}
\begin{tabular}{ccccc|cccc|cccc}
\toprule
\multicolumn{1}{l}{} & \multicolumn{4}{c|}{$s_n$ (Node)} & \multicolumn{4}{c|}{$s_e$ (Edge)} & \multicolumn{4}{c}{$s_n+s_e$ (VEGA)} \\ \cmidrule{2-13}
\multicolumn{1}{l}{} & $R_5$ & $\tau_5$ & $\tau$ & Top-1 Acc. & $R_5$ & $\tau_5$ & $\tau$ & Top-1 Acc. & $R_5$ & $\tau_5$ & $\tau$ & Top-1 Acc. \\
\midrule
(a) & 0.62 & {\color[HTML]{C00000} 0.49} & 0.60 & {\color[HTML]{C00000} 0.62} & 0.44 & -0.13 & 0.44 & 0.61 & {\color[HTML]{C00000} 0.64} & 0.20 & {\color[HTML]{C00000} 0.62} & {\color[HTML]{C00000} 0.62} \\
(b) & {\color[HTML]{C00000} 0.68} & 0.39 & 0.56 & {\color[HTML]{C00000} 0.67} & 0.68 & 0.01 & 0.01 & 0.63 & 0.64 & {\color[HTML]{C00000} 0.46} & {\color[HTML]{C00000} 0.64} & {\color[HTML]{C00000} 0.67} \\
(c) & {\color[HTML]{C00000} 0.42} & {\color[HTML]{C00000} 0.03} & 0.55 & {\color[HTML]{C00000} 0.63} & 0.16 & 0.00 & 0.55 & {\color[HTML]{C00000} 0.63} & 0.36 & -0.07 & {\color[HTML]{C00000} 0.56} & {\color[HTML]{C00000} 0.63} \\
\bottomrule
\end{tabular}
}
\label{tab:abl}
\end{table*}

\subsection{Ablation Study}

Table \ref{tab:abl} presents the ablation study of VEGA on three benchmarks mentioned in the above sections: (a) prediction on VLMs from the CLIP family (Sec. \ref{sec:main}), (b) prediction on VLMs from various pre-training algorithms (Sec. \ref{sec:mm}), and (c) prediction on combinations of VLMs and prompt templates (Sec. \ref{sec:mix}). 
The study investigates the contributions of node similarity $s_n$ and edge similarity $s_e$ individually, as well as their combination $s_n+s_e$ which constitutes the full VEGA method.
In all cases, the full VEGA method, which combines both node and edge similarity, achieves the highest predictive accuracy with $R^2$ and $\rho$ values surpassing those of using $s_n$ or $s_e$ individually. 
For the CLIP family benchmark (a), VEGA achieves the best in three of the four metrics, indicating its strong predictive capability. 
Similar trends are observed in the other two benchmarks, with VEGA outperforming its individual components, highlighting the robustness and effectiveness of integrating both node and edge similarities for downstream performance prediction.
The role of node similarity is greater than that of edge, and the combination of edge and node can improve $\tau$, indicating that the sum of node and edge similarity can more comprehensively evaluate the performance of the model.

\subsection{Sensitive Analysis}
\begin{figure}[ht]
    \centering
    \includegraphics[width=1\linewidth]{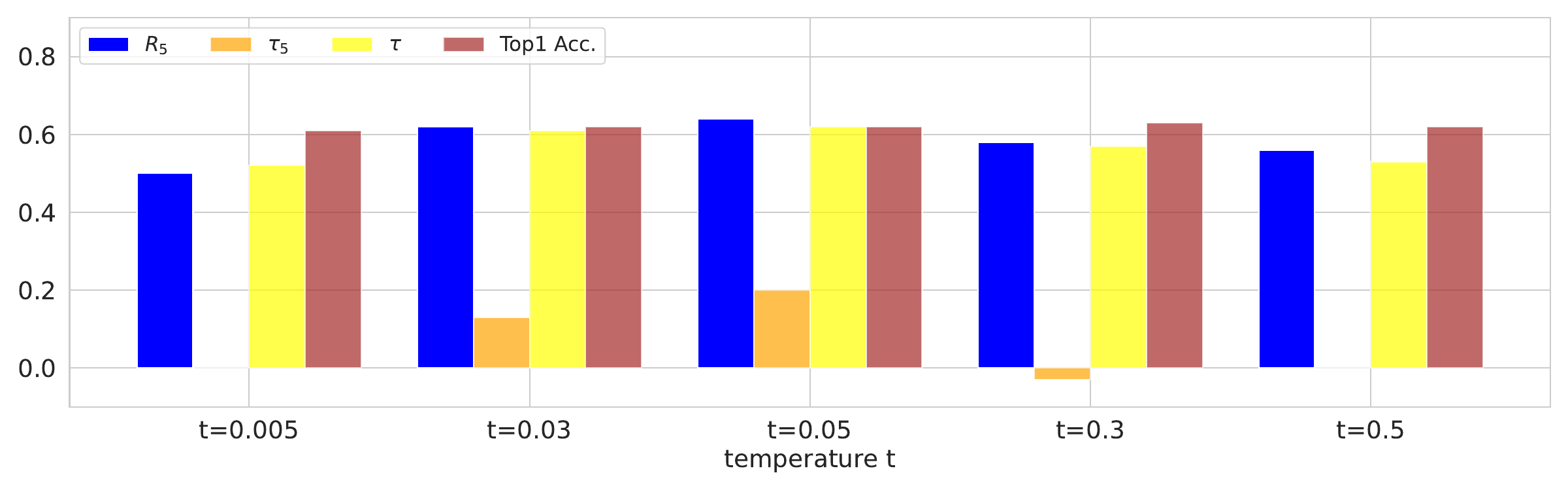}
    \caption{Sensitivity analysis on temperature $t$ in Eq. (\ref{eq:nodesim}). The Y-axis is the average results of the prediction for VLMs from CLIP family (Sec \ref{sec:main}).}
    \label{fig:temp}
\end{figure}

In the calculation of node similarity in Eq. (\ref{eq:nodesim}), we introduce a temperature parameter $t$ to sharpen the node score $s_n$.
The value of $t$ is empirically set to 0.05 and is kept consistent across all experiments, including different models and downstream datasets. 
In this section, we provide a sensitivity analysis of $t$ on the prediction for VLMs from CLIP family.
For each dataset, the figure presents the values of four metrics for five different temperature settings around the default $t$=0.05, ranging from $t$ = 0.005 to $t$ = 0.5. 
We report the average results on ten downstream datasets in Fig. \ref{fig:temp}.
The results indicate that VEGA maintains stable performance across varying temperatures.

\section{Conclusion}
This paper introduces a novel method called Visual-tExtual Graph Alignment (VEGA) for unsupervised vision language model selection, without access to downstream dataset annotations or the training data of VLMs. 
The core intuition behind VEGA is that models with similar structures in textual and visual features are more effective at matching images with their corresponding labels. 
Specifically, we construct two fully connected graphs representing the class distributions for visual and textual modalities, and define the VEGA score as the similarity between these two graphs.
We establish three benchmarks across practical application scenarios for VLM performance prediction. 
VEGA outperforms existing baselines, demonstrating its effectiveness and reliability in estimating VLM performance for unlabeled downstream tasks, and the generalizability in various scenarios.
We hope this work provides valuable insights for further research in this field.

\bibliographystyle{IEEEtran}
\bibliography{ref}

\vfill

\end{document}